\documentclass{IOS-Book-Article}
\usepackage{amsmath}
\usepackage{paralist}
\usepackage{times}
\normalfont
\usepackage[T1]{fontenc}
\usepackage{graphicx}

\newcommand{\Comment}[1]{}

\begin{document}
\begin{frontmatter}                           

\title{Compositional Verification for Autonomous Systems with Deep Learning Components}
\subtitle{White Paper}

\author{Corina S. P\u{a}s\u{a}reanu$^{*}$}
\footnote{Corresponding Author: Corina S. P\u{a}s\u{a}reanu, NASA Ames Research Center, MS 269-2, Moffett Field, CA 94035, USA; E-mail: corina.pasareanu@west.cmu.edu}
\author{Divya Gopinath$^{*}$}
\author{Huafeng Yu$^{**}$}%

\address{$^*$Carnegie Mellon University and NASA Ames, and $^{**}$Boeing}

\maketitle

\begin{abstract}
As autonomy becomes prevalent in many applications, ranging from recommendation systems to fully autonomous vehicles, there is an increased need to provide safety guarantees for such systems. The problem is difficult, as these are large, complex systems which operate in uncertain environments, requiring data-driven machine-learning components. However, learning techniques such as Deep Neural Networks, widely used today, are inherently unpredictable and lack the theoretical foundations to provide strong assurance guarantees. We present a compositional approach for the scalable, formal verification of autonomous systems that contain Deep Neural Network components. The approach uses assume-guarantee reasoning whereby {\em contracts}, encoding the input-output behavior of individual components, allow the designer to model and incorporate the behavior of the learning-enabled components working side-by-side with the other components. We illustrate the approach on an example taken from the autonomous vehicles domain.

{\bf Keywords:} model checking, compositional verification, assume-guarantee reasoning, autonomy, deep learning
\end{abstract}
\end{frontmatter}

\thispagestyle{empty}
\pagestyle{empty}


\section{Introduction}
\label{sec:intro}


Autonomy is increasingly preva;ent in many applications, ranging from
recommendation systems to fully autonomous vehicles, that require
strong safety assurance gurantees. However, this is difficult to
achieve, since autonomous sytems are large, complex systems, that
operate in uncertain environment conditions and often use data-driven,
machine-learning algorithms.  Machine-learning techniques such as deep
neural nets (DNN), widely used today, are inherently unpredictable and
lack the theoretical foundations to provide the assurance guarantees
needed by safety-critical applications. Current assurance approaches
involve design and testing procedures that are expensive and
inadequate, as they have been developed mostly for human-in-the-loop
systems and do not apply to systems with advanced autonomy.

\begin{figure}[t]
  \centering
\includegraphics[scale=0.45]{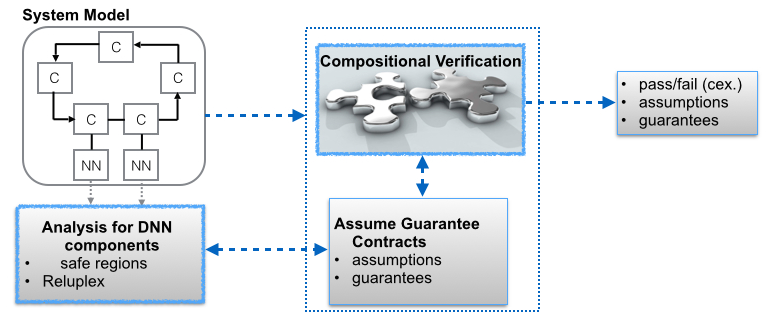}
  \caption{Overview}
  \label{fig:overview}
\end{figure}

We propose a compositional approach for the scalable verification of
learning-enabled autonomous systems to achieve design-time
assurance guarantees.
The approach is illustrated in Figure~\ref{fig:overview}. The input to
the framework is the design model of an autonomous system (this could be given 
as e.g. Simulink/Stateflow or prototype implemntation). As the
verification of the system as a whole is likely intractable we
advocate the use of compositional assume-guarantee verification
whereby formally defined {\em contracts} allow the designer to model
and reason about learning-enabled components working side-by-side with
the other components in the system. These contracts encode the
properties {\em guaranteed} by the component and the environment {\em
  assumptions} under which these guarantees hold. The framework
will then use compositional reasoning to {\em decompose} the verification of large
systems into the more manageable verification of individual components, 
which are formally checked against their respective assume-guarantee contracts.
The approach enables separate component verification with specialized tools 
(e.g. one can use software model checking for a dicrete-time controller but hybrid model checking for 
the plant component in an autonomous sytem) and seamless integration
of DNN analysis results.

For DNN analysis, we proposde to use clustering techniques to automatically
discover {\em safe regions} where the networks behave in a {\em
  predictable} way. The {\em evidence} obtained from this analysis is
{\em conditional}, subject to constraints defined by the safe regions,
and is encoded in the assume-guarantee contracts. The contracts allow
us to relate the DNN behavior to the validity of the system-level
requirements, using compositional model checking.  We illustrate the
approach on an example of an autonomous vehicle that uses DNN in the
perception module.


\section{Compositional Verification}
\label{sec:compositional}

Formal methods provide a rigorous way of obtaining strong assurance
guarantees of computing systems.  There are several challenges to
formally modeling and verifying autonomous systems. Firstly, such
systems comprise of many \emph{heterogeneous components}; each with
different implementations and requirements, which can be addressed
best with different verification models and techniques. Secondly, the
\emph{state space of such systems is very large}. Suppose we could
model all the components of such a system as formally specified
(hybrid) models; even ignoring the learning aspect, their composition
would likely be intractable. The DNN components make the scalability
problem even more serious: for example the feature space of RGB
1000X600px pictures for an image classifier used in the perception
module of an autonomous vehicle contains 256$^{1000 X 600 X 3}$
elements. Last but not the least, it is not clear how to formally
reason about the DNN components as there is no clear consensus in the
research community on a \emph{formal definition of correctness for the
  underlying machine learning algorithms}.

We propose a compositional assume-guarantee verification approach for
the scalable verification of autonomous systems where DNN components
are working side-by side with the other components. Compositional
verification frameworks have been proposed before to improve the
reliability and predictability of
CPS~\cite{DBLP:conf/emsoft/BakC16,DBLP:journals/corr/LiNSXL17,DBLP:journals/tcs/ChiltonJK14,DBLP:journals/scp/ChiltonJK14},
but none of these works address systems that include DNN
components. Recent work~\cite{DBLP:conf/nfm/DreossiDS17} proposes a
compositional framework for the the analysis of autonomous systems
with DNN components. However, that approach addresses {\em
  falsification} in such systems and, while that is very useful for
debugging, it is not clear how it can be used to provide assurance
{\em guarantees}.

Assume-guarantee reasoning attempts to break up the verification of a large
system  into the local verification of individual components, using {\em assumptions} about
the rest of the system.
The simplest assume-guarantee rule first checks that a component $M_1$
satisfies a property $P$ under an assumption $A$ (this can be written
as $M_1\models A\rightarrow P$). If the ``environment" $M_2$ of $M_1$
(i.e., the rest of the system in which $M_1$ operates) satisfies $A$
(written as $M_2\models true \rightarrow P$), then we can prove that the
whole system composed of $M_1$ and $M_2$ satisfies $P$.
Thus we can decompose the global property $P$ into two local assume-guarantee
properties (i.e., contracts) $A\rightarrow P$ and $A$ that are expected
to hold on $M_1$ and $M_2$ respectively.
Other, more involved, rules allow reasoning about the circular
dependencies between components, where the assumption for one
component is used as the guarantee of the other component and vice
versa; if the conjunction of the assumptions implies the specification
than the overall system guarantees the system-level requirement.
Rules that involve circular reasoning use inductive arguments, over
time, formulas to be checked, or both, to ensure soundness.
Furthermore, the rules can be naturally generalized to reasoning about
more than two components and use different notions for property
satisfaction such as trace inclusion or refinement checking.

The main challenge with assume-guarantee reasoning techniques is to come up with assumptions
and guarantees that can be suitably used in the assume-guarantee rules.
This is typically a difficult manual process.
Progress has been made on automating assume-guarantee reasoning using
learning and abstraction-refinement techniques for iterative building
of the necessary assumptions~\cite{DBLP:journals/fmsd/PasareanuGBCB08}.
The original work was done in the context of systems expressed
as finite-state automata, but progress has been made in the automated compositional verification for probabilistic and hybrid systems~\cite{DBLP:conf/cav/KomuravelliPC12,DBLP:conf/hvc/BogomolovFGGPPS14}, 
which can be used to model autonomous systems.

Assume-guarantee reasoning can be used for the verification of
autonomous systems either by replacing the component with its
assume-guarantee specification in the compositional proofs or by using
an assume-guarantee rule such as the above to decompose the
verification of the systems into the verification of its
components. Furthermore, the assume-guarantee specifications can be
used to drive component-based testing and run-time monitoring, in the
cases where the design-time formal analysis is not possible, either
because the components are too large or they are {\em adaptive},
i.e. the component behavior changes at run-time (using
e.g. reinforcement learning).

\section{Analysis for Deep Neural Network Components}
\label{sec:DNN}

Deep neural networks (DNNs) are computing systems inspired by the biological neural networks that constitute animal brains. They consist of neurons (i.e. computational units) organized in many layers. These systems are capable of {\em learning} various tasks from {\em labeled examples} without requiring task-specific programming. DNNs have achieved impressive results in computer vision, autonomous transport, speech recognition, social network filtering, bioinformatics and many other domains and there is increased interest in using them in safety-critical applications that require strong assurance guarantees. However, it is difficult to provide such guarantees since it is known that these networks can be easily fooled by \empty{adversarial perturbations}: minimal changes to correctly-classified inputs, that
cause the network to misclassify them. For instance, in image-recognition networks it is possible to
add a small amount of noise (undetectable by the human eye) to an
image and change how it is classified by the network. 

This phenomenon represents a
safety concern, but it is currently unclear how to measure a
network's robustness against it.
To date, researchers have mostly focused on efficiently
finding adversarial perturbations around select individual input points. The goal is to find an input
$x'$ as close as possible to a known input $x$ such that $x'$ and $x$ are labeled
differently. Finding the optimal solution for this problem is
computationally difficult, and so various approximation approaches
have been proposed. Some approaches are \emph{gradient based}~\cite{SzegedyZSBEGF13,GoodfellowSS14,FeinmanCSG17},
whereas other use optimization techniques~\cite{Carlini017}. These approaches have successfully demonstrated the weakness of many
state-of-the-art networks; however, these approaches operate on individual
input points, and it is unclear how to apply them to large input domains, unless one does a brute-force enumeration of all input values which is infeasible for most practical purposes. Furthermore, because they are inherently incomplete, these techniques can not even provide any guarantees around the few selected individual points. Recent approaches tackle neural network verification~\cite{HuangKWW17,KaBaDiJuKo17Reluplex}
by casting it as an SMT solving problem.  Still, these techniques operate best when applied to individual points and further do not have a well-defined rationale to select meaningful regions around inputs within which the network is expected to behave consistently.

In ~\cite{DEEPSAFE}, we developed a DNN analysis  to
 automatically discover {\em input regions} that are likely to be
 robust to adversarial perturbations, i.e. to have the same true
 label, akin to finding likely invariants in program analysis. The
 technique takes inputs with known true labels from the training set
 and it iteratively applies a clustering
 algorithm~\cite{KanungoMNPSW02} to obtain small groups of inputs that
 are close to each other (with respect to different distance metrics)
 and share the same true label. Each cluster defines a {\em region} in
 the input space (characterized by the centroid and radius of the
 cluster).  Our hypothesis is that for regions formed from dense
 clusters, the DNN is well-trained and we expect that all the other
 inputs in the region (not just the training inputs) should have the
 same true label.  We formulate this as a safety check and we verify
 it using off-the-shelf solvers such as
 Reluplex~\cite{KaBaDiJuKo17Reluplex}.
If a region is found to be {\em safe}, we
provide \emph{guarantees w.r.t all points within that region}, not
just for individual points as in previous techniques.

As the usual notion of safety might be too strong for many DNNs, we
introduce the concept of \emph{targeted safety}, analogous to
targeted adversarial perturbations \cite{SzegedyZSBEGF13,GoodfellowSS14,FeinmanCSG17}. The
verification checks \emph{targeted safety} which, given a specific
incorrect label, guarantees that no input in the region is mapped by
the DNN to that label. Therefore, even if in that region the DNN is not
completely robust against adversarial perturbations, we give
guarantees that it is safe against specific targeted attacks.

As an example, consider a DNN used for perception in an autonomous car
that classifies the images of a semaphore as red, green or yellow. We
may want to guarantee that the DNN will never classify the image of a
green light as a red light and vice versa but it may be tolerable to
misclassify a green light as yellow, while still avoiding traffic
violations.

The safe regions discovered by our technique enable characterizing the input-output behavior of the network over partitions of the input space, which can be encoded in the assume-guarantee specifications for the DNN components.
The regions will define the conditions (assumptions), and the guarantees will be that all the points within the region will be assigned the same labels. The regions could be characterized as geometric shapes in Euclidean space with centroids and radii. The conditions would then be in terms of standard distance metric constraints on the input attributes. For instance, all inputs within a Euclidean distance $r$ from the centroid $cen$ of the region would be labeled $l$ by the network.

Note that the verification of even simple neural networks is an NP-complete problem and is very difficult in practice. Focusing on clusters means that verification can be applied to small input domains, making it more feasible and rendering the approach as a whole more scalable. Further, the verification of separate clusters can be done in parallel, increasing scalability even further.

In~\cite{DEEPSAFE} we applied the technique on the MNIST dataset ~\cite{MNIST} and on a
neural network implementation of a controller for the next-generation Airborne
Collision Avoidance System for unmanned aircraft (ACAS Xu)~\cite{ACASXU}, where we
used Reluplex for the safety checks.
For these networks,
our approach identified multiple regions which were completely safe as well as
some which were only safe for specific labels.
It also discovered adversarial examples  which were confirmed by domain experts.
We discuss the ACAS Xu experiments in more detail below.

\subsection{ACAS Xu case study}

\begin{figure}[t]
\centering
	\includegraphics[scale=0.35]{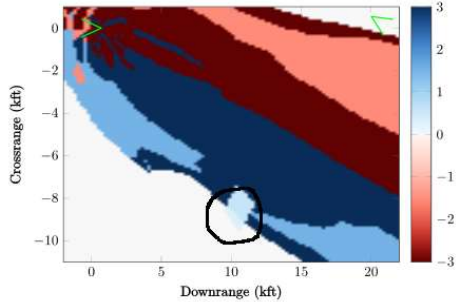}\vspace{-3cm}
    \caption{Inputs highlighted in light blue are mis-classified as
      Strong Right instead of COC. $\text{Crossrange}= \rho
      \cdot \sin(\theta)$, $\text{Downrange} = \rho \cdot \cos(\theta)$.\label{fig:adv3}}
\end{figure}

ACAS X is a family of collision avoidance systems for aircraft which
is currently under development by the Federal Aviation Administration (FAA)~\cite{ACASXU}.
ACAS Xu is the version for unmanned aircraft control. It is intended
to be airborne and receive sensor information regarding the drone (the

\emph{ownship}) and
any nearby intruder drones, and then issue horizontal turning
advisories aimed at preventing collisions. The input sensor data includes:

\begin{itemize}
  \item $\rho$: distance from ownship to intruder;
  \item $\theta$: angle of intruder relative to ownship heading direction;
  \item $\psi$: heading angle of intruder relative to ownship heading direction;
  \item $v_{own}$: speed of ownship;
  \item $v_{int}$: speed of intruder;
  \item $\tau$: time until loss of vertical separation; and
  \item $a_{prev}$: previous advisory.
\end{itemize}

The five possible output
actions are as follows: Clear-of-Conflict (COC), Weak Right, Weak
Left, Strong Right, and Strong Left. Each advisory is assigned a score,
with the lowest score corresponding to the best action. The FAA is currently exploring an implementation of ACAS Xu that uses
an array of 45 deep neural networks. These networks were obtained by
discretizing the two parameters, $\tau$ and $a_{prev}$, and
so each network contains five input dimensions and treats
$\tau$ and $a_{prev}$ as constants. Each network
has 6 hidden layers and a total of 300 hidden ReLU activation
nodes. We were supplied a set of cut-points, representing valid important values for each dimension, by the domain experts~\cite{ACASXU}. We generated a set of 2662704 inputs (cartesian product of the values for all the dimensions). The network was executed on these inputs and the output advisories (labels) were verified. These were considered as the inputs with known labels for our experiments.

We were able to prove safety for 177 regions in total (125 regions where the network was completely safe against mis-classification to any label and 52 regions where the network was safe against specific target labels). An example of the safety guarantee is as follows;
\begin{equation} \label{eq:com-safe-region}
  \forall \  x \in \ | x - \{0.19,0.31,0.28,0.33,0.33\} |_{L_1} \leq 0.28 \quad\Rightarrow\quad label(x) = COC
\end{equation}
Here \{0.19,0.31,0.28,0.33,0.33\} are the normalized values for the 5 input attributes ($\rho$,$\theta$,$\psi$,$v_{own}$,$v_{int}$ ) corresponding to the centroid of the region  and 0.28 is the radius. The distance is in the Manhattan distance metric (L1). The contract states that under the condition that an input lies within 0.28 distance from the input vector \{0.19,0.31,0.28,0.33,0.33\}, the network is guaranteed to mark the action for it as COC which is the desired output.



Our analysis also discovered adversarial examples of interest, which were validated by the developers. 
Fig.~\ref{fig:adv3} illustrates such an example for ACAS Xu.

The safety contracts obtained with the region analysis can be used in the compositional verification of the overall autonomous systems, which can be performed with standard model checkers.

\section{Example}
\label{sec:example}

\begin{figure}
  \centering
  \includegraphics[scale=0.55]{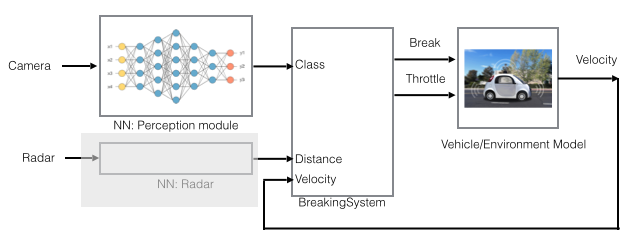}
  \caption{Example}
  \label{fig:ex}
\end{figure}

We illustrate our compositional approach on an example of an autonomous vehicle.
The platform includes learning components that allow it to detect other
vehicles and drive according to traffic regulations; the platform also
includes reinforcement learning components to evolve and refine its
behavior in order to learn how to avoid obstacles in a new
environment.

We focus on a subsystem, namely an automatic emergency breaking system, illustrated in
Figure~\ref{fig:ex}. It has three components: the \emph{BreakingSystem},
the \emph{Vehicle} (which, to simplify the presentation, we assume it
includes both the autonomous vehicle and the environment) and a
\emph{perception module} implemented with a DNN; there may be
other sensors (radar, LIDAR, GPS) that we abstract away here for
simplicity. The breaking system sends signals to the vehicle to regulate the
acceleration and breaking, based on vehicle velocity, distance to obstacles
 and traffic signals. The velocity information is
provided as a feedback from the plant, the distance information is
obtained from sensors, while the information about traffic lights is
obtained from the perception module. The perception module acts as a
classifier over images captured with a camera. Such systems are
already employed today in semi-autonomous vehicles where adaptive
cruise controllers or lane keeping assist systems rely on image
classifiers providing input to the software controlling electrical and
mechanical subsystems~\cite{DBLP:conf/nfm/DreossiDS17}.
Suppose we want to check that the system satisfies the following
\emph{safety property}: the vehicle will not enter an intersection if the
traffic light at the intersection turns red.

We write the system as the composition: $\emph{BreakingSystem} || \emph{Vehicle}
|| \emph{NN}$. Each component has an interface that specifies its input and output variables (ports), and their parallel composition is formed
by connecting components via ports. We write the property as follows
(using Linear Temporal Logic, LTL, assuming discrete time): globally (G) if the semaphore (input image {\em x}) is
red then eventually (F), within 3 seconds, the velocity becomes 0:

$$P::\\\quad G ((x=red) \quad\Rightarrow\quad F_{T<3s} (velocity=0))$$

In practice, we would also need to encode in $P$ the
assumption that the distance to traffic light is less than some
threshold, but we simplify here to ease the presentation.  We are thus
interested in checking that the system satisfies
property $P$, written as $S \models P$.
We decompose the system into two subsystems:
$M_1=\emph{BreakingSystem} ||  \emph{Vehicle}$ and $M_2=\emph{NN}$ and define two assume-guarantee contracts $C_1$ and $C_2$ for the two subsystems.
Suppose (part of) the contract for $M_1$ is:

$$C_1::\\\quad G ((Class=red) \quad\Rightarrow\quad  F_{T<3s} (velocity=0))$$

The contract states that {\em assuming} the input (Class) to the subsystem $M_1$ is red then the vehicle is {\em guaranteed} to stop in at most 3 time units.
We can further decompose the verification of $M_1$ into the separate verification of its components using additional contracts and perform component-wise verification. It remains to formally characterize the input-output behavior of the DNN in a contract that can be used in the compositional proofs. This is a difficult problem because DNN are known to be vulnerable to adversarial perturbations~\cite{SzegedyZSBEGF13,KuGoBe16}: a small perturbation added to an image that shows a red semaphore might lead the NN misclassifying it as having $Class=green$.

To address the problem, we use clustering over the training set (see Section~\ref{sec:DNN}) to automatically find regions where the network is likely to be robust to adversarial perturbations.
The result is a finite set $\mathcal{R}$ of well-defined regions, where a region
$\rho\in \mathcal{R}$ is characterized by a pair $(c,r)$;
$c$ is the centroid and $r$ is the radius of the region.
We then use a verification tool (such as Reluplex) to check that, {\em for all inputs} $x$ within each region, the NN classifies them to the same label as that of known inputs (and of $c$):


$$C_{\rho}::\\\quad|x-c|<r \  \quad\Rightarrow\quad \ Label(x)=Label(c)$$

The training data available and the amount of noise could impact the validity of the check.
In such cases we may need to refine the contracts to include Bayesian estimates of uncertainty~\cite{DROPNN}. 
Let $Uncert(x)$ denote the uncertainty in the output of the NN for an input $x$.
We can then refine the contract to check that the label is as expected {\em and} the uncertainty level is below a threshold.
The DNN's {\em safety contract} $C_2$ could then be the union of all the constraints of the form $C_{\rho}$ that are proved valid.


We are now ready to perform the compositional proof: if $M_1 \models C_1$ and $M_2 \models C_2$ and furthermore $C_1 \wedge C_2 \Rightarrow P$, it follows that $M_1 || M_2 \models P$; thus we prove that the whole system satisfies the property, without composing its (large) state space. This proof can be performed with standard model checkers.
\Comment{
$\begin{array}{ll}
1. & BreakingSystem || Vehicle \models C_1\\
2. & NN \models C_2\\
3. & C_1 \wedge C_2 \quad\Rightarrow\quad P\\
\hline
& Controller || Plant || NN \models P.
\end{array}$

Thus if all the premises are satisfied, we can prove that the system satisfies the property without composing the state space of the whole system.
}

\subsection{Run-time Monitoring and Control}
We note that the evidence we obtain from the analysis is {\em conditional};
we can only prove that the property holds for the region contracts that we found to be safe.
The information encoded in the contract assumptions will need to be used to
synthesize {\em run-time guards} that monitor inputs that fall outside the conditions and instruct the system to take appropriate, fail-safe actions.
Note also that this compositional approach enables separate verification of individual components: we can thus replace some of the verification tasks for individual components with testing or simulation, which will increase scalability but will give only empirical guarantees. 

Furthermore, if the system contains {\em adaptive} components, the verification of those components can be done at runtime, whereas the static components only need to be checked once, at design time.  Adaptive learning-enabled components pose additional challenges over time. We can again use model uncertainty to identify situations in which the adaptrive learning-enabled system is not confident about its decisions, and take appropriate actions in such cases.



\section{Conclusion}
\label{sec:conclusion}

We presented a compositional approach for the verification of autonomous systems. The approach uses assume-guarantee reasoning for scalable verification and can naturally integrate reasoning about the learning-enabled components in the system. We are working on evaluating the proposed approach on various simulation and real autonomous platforms, including self-driving cars (discussed briefly in Section~\ref{sec:example}), autonomous quadcopters and airplanes. These case studies cover  perception, decision making, control and actuation of autonomous systems, and they include safety-critical cyber-physical components as well as DNN components. 

\bibliographystyle{abbrv}
\bibliography{references,Refs}


\end{document}